\documentclass[twoside ]{article}
\usepackage{cite}
\usepackage{graphicx}
\usepackage{subfloat}
\RequirePackage{epsfig}

\usepackage{amsmath}
\usepackage{amssymb}
\usepackage{verbatim} 
\usepackage{algorithm}
\usepackage{listings}
\usepackage{textcomp}
\usepackage{relsize}
\usepackage{epstopdf}

\newtheorem{theorem}{Theorem}[section]

\newtheorem{proposition}[theorem]{Proposition}

\newcommand{\qed}{\nobreak \ifvmode \relax \else
      \ifdim\lastskip<1.5em \hskip-\lastskip
      \hskip1.5em plus0em minus0.5em \fi \nobreak
      \vrule height0.75em width0.5em depth0.25em\fi}
\usepackage[accepted]{aistats2014}
\begin{document}

\twocolumn[

\aistatstitle{To go deep or wide in learning?}

\aistatsauthor{ Gaurav Pandey and Ambedkar Dukkipati}

\aistatsaddress{ Department of Computer Science and Automation \\ Indian Institute of Science, Bangalore 560012, India \\  }
]





\begin{abstract}
  To achieve acceptable performance for AI tasks, one can either use sophisticated feature extraction methods as the first layer in a two-layered supervised learning model, or learn the features directly using a deep (multi-layered) model. While the first approach is very problem-specific, the second approach has computational overheads in learning multiple layers and fine-tuning of the model. In this paper, we propose an approach called wide learning based on arc-cosine kernels, that learns a single layer of infinite width. We propose exact and inexact learning strategies for wide learning and show that wide learning with single layer outperforms single layer as well as deep architectures of finite width for some benchmark datasets.
\end{abstract}

\section{INTRODUCTION}

In spite of the vast research on machine learning and AI in the past many decades, we still do not have systems that can perform as well as humans for many real world tasks. For instance, for the task of scene labelling, human performance far exceeds the performance of all known learning algorithms~\cite{xiao2010sun}. One reason often cited for this difference in performance is the insufficient depth of the architecture used by the learning algorithms~\cite{bengio2009learning}. Typically, supervised learning algorithms can be thought to have 2 layers (i.e., depth = 2), whereby, in the first layer, the data is mapped to a feature space (either using the kernel trick or by feature extraction~\cite{lowe2004distinctive, dalal2005histograms}), followed by learning a linear classifier in the feature space. Though the aim is to map the data into a feature space, where the classes become linearly separable, these two stages of learning are often kept independent of each other, with notable exceptions being metric learning~\cite{blitzer2005distance} and multiple kernel learning~\cite{bach2004multiple}. Such architectures are called shallow architectures~\cite{bengio2009learning}.

In recent years, architectures with multiple layers have been shown to achieve state-of-the-art results for many AI tasks such as image and speech recognition, often outperforming other shallow architectures by a huge margin~\cite{ciresan2012multi, wan2013regularization}. Most of these architectures are minor modifications of multi-layer neural networks (MLNN). Though MLNN have been around for more than 2 decades, architectures with more than 2 hidden layers were seldom used for learning~\cite{utgoff2002many}. It was argued that multi-layered neural networks get stuck in bad local optima, thereby giving bad generalization performance. This trend was broken in 2006, when Hinton et al.~\cite{hinton2006fast} showed that a multi-layered neural network can be effectively trained to achieve good generalization performance, if the weights of connections between various hidden layers were initialized by using the data only and not their labels in a purely unsupervised fashion using restricted Boltzmann machines~(RBMs). He called such models as deep belief nets (DBN). 

The success of deep learning methods over shallow architectures have caused the revival of multi-layer neural networks in machine learning literature. For most AI tasks, multi-layer neural networks (that may be initialized with unsupervised pre-training) tend to outperform most other learning algorithms. Unfortunately, the algorithms used for multi-layered neural networks are often non-convex and rely on a number of heuristics, such as number of epochs, learning rate, momentum, batch size etc. Incorrect tuning of these hyper-parameters often results in decrease in performance~\cite{bergstra2011algorithms}. Generally, computationally costly grid searches are used to get a suitable value for the hyper-parameters. Furthermore, the algorithms themselves are computationally complex and often rely on the processing power of GPUs~\cite{ciresan2012multi, krizhevsky2012imagenet}.

Due to the above mentioned deficiencies in neural networks, one may wonder whether one can achieve the same performance as in deep learning without ever resorting to neural networks? Ideally, a good feature learning algorithm should be able to learn features from the data that disentangle the factors of variation~\cite{bengio2009learning}. In such a case, no fine-tuning of the model using multi-layer neural networks would be necessary. In other words, we can directly take the learnt features as input and feed it to a classifier to get the labels as output. However, results tend to suggest that models that have not been fine-tuned tend to perform poorly. This has led many researchers to believe that restricted Boltzmann machines~(RBMs) are not doing a good job at capturing the invariances and disentangling the underlying factors of variation present in data~\cite{lamblin2010important}.

It is in this light that we present our paper. We introduce wide-learning using covariance arc-cosine kernels (based on arc-cosine kernels introduced in~\cite{cho2010large}), that builds upon the existing RBM model. We give exact as well as non-exact methods for learning in such models. In particular, we show that output of the RBM can be directly used to learn the covariance matrix in a covariance arc-cosine kernel. We show that the model is actually capable of learning features, that makes the fine-tuning stage redundant. Secondly, we also show that for the datasets considered, a single wide layer~(that is, a single layer of covariance arc-cosine kernel) is sufficient to generate a representation that helps to capture invariances in the data, since preliminary results suggest that stacking multiple wide layers leads to decrease in performance. Using a single RBM to learn a wide layer, we are able to obtain better results for many classification tasks than obtained by multi-layer neural network initialized using a deep belief network and fine-tuned using backpropagation.

\section{PRELIMINARIES AND BACKGROUND}
\subsection{Restricted Boltzmann Machine (RBM)}
An RBM~\cite{hinton2002training} is a complete bipartite Markov random field with a layer of visible units $(\mathbf{x})$ and another layer of finitely many latent units $(\mathbf{h})$. The visible units correspond to the features of the observed sample, for instance, pixels in an image. Every visible unit is connected to every hidden unit by an edge. Since the graph is bipartite, the cliques of the model correspond to the edges and have size $2$. The potential function of an edge $(v_i, h_j)$ is given by $-(w_{ij}v_ih_j + a_iv_i + b_jh_j)$, where $w_{ij}, a_i, b_j, \: 1\le i \le d, \: 1 \le j \le K$ from the parameters of the model. The energy function which is the sum of potential function across all edges, is given by
\begin{align}
E(\mathbf{x},\mathbf{h}) &= - \sum_{i,j} w_{ij}v_ih_j + a_iv_i + b_jh_j \\
						 &= - (\mathbf{x}^TW\mathbf{h} + a^T\mathbf{x} + b^T\mathbf{h})
\end{align}
The corresponding marginal probability of an instance $\mathbf{x}$ is given by
\begin{equation}
p(\mathbf{x}) = \frac{\sum_{\mathbf{h}} \exp(-E(\mathbf{x},\mathbf{h}) )}{\sum_{\mathbf{x}', \mathbf{h} } \exp(-E(\mathbf{x}',\mathbf{h}) )}\:. \label{stochasticRBM}
\end{equation}

In order to maximize the log-likelihood of an RBM for a sequence of observations, one can use stochastic gradient descent techniques. The gradient of the log-likelihood for a fixed observation $\mathbf{x}$ with respect to the weight matrix $W$ is given by
\begin{align}
\nabla_{W}L &= \frac{1}{N} \sum_{n=1}^N \mathbb{E}_{p(\mathbf{h}|\mathbf{x}_{(n)})} \mathbf{x}_{(n)}\mathbf{h}^T - \mathbb{E}_{p(\mathbf{x},\mathbf{h})} \mathbf{x}\mathbf{h}^T \: . \label{Wupdate}
\end{align}
The gradient with respect to other parameters can be computed similarly. The first quantity in the RHS is straightforward to compute from the following equations.
\begin{align}
\mathbb{E}_{p(\mathbf{h}|\mathbf{x})} h_j &= \frac{\sum_{\mathbf{h}}\exp(\mathbf{x}^TW\mathbf{h} + a^T\mathbf{x} + b^T\mathbf{h}) h_j}{\sum_{\mathbf{h}}\exp(\mathbf{x}^TW\mathbf{h} + a^T\mathbf{x} + b^T\mathbf{h}))} \\
&= \frac{1}{1+\exp(-(\mathbf{x}^Tw_j + b_j))} \:,
\end{align}
where $w_j$ is the $k^{th}$ column in $W$.
In order to compute the second quantity, we need the expected value of $\mathbf{x}\mathbf{h}^T$ for the current choice of $W$. This can be obtained by using Gibbs sampling. In practise, a small number~(p) of iterations of Gibbs sampling is run to get $\mathbf{x}_p$ and $\mathbf{h}_p$ and plugged in equation~\eqref{Wupdate}. This method, also known as Contrastive Divergence (CD)~\cite{hinton2002training}, has been shown to give a good approximation to the actual gradient. The corresponding update equation is given by
\begin{equation}
W^{t+1} = W^{t} + \eta(\mathbf{x}_0\mathbf{h}_0^T - \mathbf{x}_p\mathbf{h}_p^T) \: ,\label{W_update}
\end{equation}
where $\eta$ is the learning rate. For more details about training an RBM,  we encourage the reader to refer to Fischer et al.~\cite{fischer2012introduction}.

A commonly used variant of RBM has rectified linear units rathar than stochastic binary units. Training of RBMs with rectified linear units is very similar to that with stochastic binary units~\cite{nair2010rectified}. The update equation for an RBM with rectified linear units is exactly similar to that of an RBM with stochastic binary units except that the hidden units are sampled uniformly from the normal distribution  with mean $\max(0, W^T\mathbf{x})$ and identity covariance matrix. In the rest of the paper, we will refer both these Markov random fields as RBM.

\subsection{Single Layer Threshold Networks And Corresponding Kernels}
In deep learning, the weights of a multi-layered neural network~(excluding those that connect the output units) are initialized using RBMs, and fine-tuned using backpropagation algorithm. Rather than pre-training the weights using RBM, it is possible to sample the weights randomly from a fixed distribution and feed the output of the hidden units directly to a linear classifier such as SVM. Contrary to intuition, it has been observed that when the weights have been sampled from standard normal distribution and the number of hidden units is much greater than the number of visible units, the resultant classifier gives good performance on many classification tasks~\cite{huang2004extreme}. Furthermore, the performance improves as the number of hidden units increase.

It is possible to perform learning tractably, when the number of hidden units in a randomly weighted neural networks tend to $\infty$ by using the kernel trick. Cho et al.~\cite{cho2010large} showed that for threshold neural networks, inner products of the randomly sampled hidden units for two instances becomes deterministic as the number of hidden units tend to $\infty$. The corresponding kernels are termed as arc-cosine kernels. Hence, learning a linear classifier in the original infinite dimensional space is same as learning a kernel machine using the arc-cosine kernel.

In particular, when the hidden units are given by
\begin{equation}
h(\mathbf{x}) = H(w^T\mathbf{x})(w^T \mathbf{x})^n, \:\:\:\:\:\: w\sim \mathcal{N}(0,1)
\end{equation} 
where $H$ is the Heavyside step function, the corresponding kernel~\cite{cho2010large} is given by
\begin{equation}
K_n(\mathbf{x}, \mathbf{y}) = \frac{1}{2\pi}\|\mathbf{x}\|^n \|\mathbf{y}\|^n J_n(\theta) \:\:, \label{arcCosineKernel}
\end{equation}
where $\theta$ is the angle between $\mathbf{x}$ and $\mathbf{y}$ and 	$J_n{\theta}$ is given by
\begin{equation}
J_n(\theta) = (-1)^n(\sin\theta)^{2n+1} \left(\frac{1}{\sin\theta}\frac{\partial}{\partial\theta}\right)^n \left(\frac{\pi-\theta}{\sin\theta}\right), \:\: n\in \mathbb{N}
\end{equation}

As a spacial case, when $n=1$, the hidden units are termed as rectified linear units and the corresponding kernel function is given by
\begin{equation}
K_1(\mathbf{x}, \mathbf{y}) = \frac{1}{2\pi}\|\mathbf{x}\|\|\mathbf{y}\|(\sin\theta + (\pi-\theta)\cos\theta) \label{rectified_linear} \enspace .
\end{equation}

\section{COVARIANCE ARC-COSINE KERNELS}
Instead of sampling the entries of the matrix $W$ from standard normal distribution, if we sample the columns from a multivariate Gaussian distribution with zero mean and covariance $\Sigma$, we get a modified arc-cosine kernel of the form 

\begin{align*}
K_{\Sigma,n}(\mathbf{x}, \mathbf{y}) = \frac{1}{(2\pi)^{\frac{d}{2}}|\Sigma|^{\frac{1}{2}}}\int_{w \in \mathbb{R}^d} H(w^T\mathbf{x}) H(w^T\mathbf{y})(w^T\mathbf{x})^n ..\\
..(w^T\mathbf{y})^n \exp \left( -\frac{w^T \Sigma^{-1} w}{2}  \right) \mathrm{d}w \:\:,
\end{align*}
which we term as covariance arc-cosine kernel.
Applying a change of variables $u = \Sigma^{-\frac{1}{2}}w$ in the above equation, we get
\begin{align*}
K_{\Sigma,n}(\mathbf{x}, \mathbf{y}) &= \frac{|\Sigma|^{\frac{1}{2}}}{(2\pi)^{\frac{d}{2}}|\Sigma|^{\frac{1}{2}}} \int_{u \in \mathbb{R}^d} H(u^T\Sigma^{\frac{1}{2}}\mathbf{x}) H(u^T\Sigma^{\frac{1}{2}}\mathbf{y})..\\&..(u^T\Sigma^{\frac{1}{2}}\mathbf{x})^n (u^T\Sigma^{\frac{1}{2}}\mathbf{y})^n \exp \left( -\frac{\|u\|^2}{2}  \right) \mathrm{d}w \:, \\
& = \frac{1}{(2\pi)^{\frac{d}{2}}}\int_{u \in \mathbb{R}^d} H(u^T\mathbf{a})H(u^T\mathbf{b}).. \\&.. (u^T\mathbf{a})^n(u^T\mathbf{b})^n \exp\left(-\frac{\|u\|^2}{2}\right)\: \mathrm{d}u \:,\\
& = K_n(\mathbf{a}, \mathbf{b})\:, \label{CovarianceKernel}
\end{align*}
where $\mathbf{a} = \Sigma^{\frac{1}{2}}\mathbf{x}$ and $\mathbf{b} = \Sigma^{\frac{1}{2}}\mathbf{y}$ respectively.
We state the above result below:
\begin{proposition}
Let the columns of the matrix $W$ be sampled from a multivariate normal distribution $\mathcal{N}(0, \Sigma)$, and let $h(\mathbf{x})= H(W^T\mathbf{x})(W^T\mathbf{x})^n$ be the representation of $\mathbf{x}$ in the feature space. Here both the Heavyside step function and the polynomial function is applied pointwise on the vector. As the number of columns in the weight matrix $W$ tend to infinity, the inner product between the feature representation is given by
\begin{equation}
K_{\Sigma,n}(\mathbf{x}, \mathbf{y}) = K_n(\mathbf{a}, \mathbf{b})
\end{equation}
where $\mathbf{a} = \Sigma^{\frac{1}{2}}\mathbf{x}$ and $\mathbf{b} = \Sigma^{\frac{1}{2}}\mathbf{y}$ respectively and $K_n$ is defined as in \eqref{arcCosineKernel}.
\end{proposition}

\section{WIDE LEARNING}
It is known~\cite{bengio2009learning} that in case of natural image patches, the features learnt by an RBM are Gabor-like, that is, they correspond to the output of a Gabor filter with some fixed frequency and orientation. Since the set of all possible frequencies and orientations has uncountably many elements, an RBM tries to extract a subset of these frequencies/orientations that best capture the invariances present in the data. A covariance arc-cosine kernel, on the other hand, tries to find a distribution over the weight vectors that best capture the invariances in the data, thereby allowing one to use infinitely many features for any given data. This is also the reason why we call distribution learning for arc-cosine kernels as wide learning.

In the rest of the paper, whenever we refer to an arc-cosine kernel, we imply the kernel with rectified linear units, that is, the kernel corresponding to $n=1$ given by~\eqref{rectified_linear}
\subsection{Exact Wide Learning}

In order to derive an algorithm for training the kernel to learn the covariance matrix $\Sigma$, we rewrite the update equation for rectified linear units mentioned in equation~\eqref{W_update} as
\begin{equation}
W^{t+1} = W^{t} + \eta(\mathbf{x}_0\mathbf{h}_0^T - \mathbf{x}_p\mathbf{h}_p^T) \enspace ,\label{W_update2}
\end{equation}
where $\mathbf{x}_0$ and $\mathbf{h}_0$ are the initial values for the visible and hidden units and $\mathbf{x}_p$ and $\mathbf{h}_p$ are the values for visible and hidden units after $p$ iteration of Gibbs sampling. In our case, we assume that $W^{t}$ has infinitely many columns sampled from some distribution with covariance matrix $\Sigma^{t}$, and we are interested in computing the covariance matrix $\Sigma^{t+1}$ of the columns of $W^{t+1}$, which is given by
 \begin{align}
 \Sigma^{t+1} &= \lim_{M\rightarrow \infty} \frac{1}{M}\sum_{k=1}^M w_k^{t+1} {w_k^{t+1}}^T \\
 		&= \lim_{M\rightarrow \infty} \frac{1}{M}W^{t+1}{W^{t+1}}^T \:\:,\label{Sigma_W}
 \end{align}
 Hence, if we multiply equation~\eqref{W_update2} by its transpose and use \eqref{Sigma_W} in the resultant equation, we get
 \begin{equation}
 \begin{split}
 \Sigma^{t+1} = \Sigma^t + &\lim_{M\rightarrow \infty} \frac{1}{M}\left[ \eta(W^{t}\mathbf{h}_0\mathbf{x}_0^T - W^t\mathbf{h_p}\mathbf{x}_p^T))	\right] \\
 		+&  \lim_{M\rightarrow \infty} \frac{1}{M}\left[ \eta(\mathbf{x}_0\mathbf{h}_0^T{W^t}^T - \mathbf{x_p}\mathbf{h}_p^T{W^t}^T))	\right] \\
 		+& \lim_{M\rightarrow \infty} \frac{1}{M} \left[ \eta^2(\mathbf{x}_0\mathbf{h}_0^T\mathbf{h}_0\mathbf{x}_0^T + \mathbf{x}_p\mathbf{h}_p^T\mathbf{h}_p\mathbf{x}_p^T - \right. \\ 
 		& \left.\:\:\:\:\:\:\:\:\:\:\:\:\:\:\:\:\:\:-\mathbf{x}_{0}\mathbf{h}_0^T\mathbf{h}_p^T\mathbf{x}_p^T - \mathbf{x}_{p}\mathbf{h}_p^T\mathbf{h}_0^T\mathbf{x}_0^T )\right] \label{Sigma_update} \enspace ,
 \end{split}
 \end{equation}
 If we assume that the $k^{th}$ unit in the hidden layer is sampled from a normal distribution with unit variance and mean $(\mathbf{x}^Tw_k)_+$~(as is commonly done for training RBMs with rectified linear units~\cite{nair2010rectified}), then
 \begin{align*}
	&\lim_{M\rightarrow \infty} 
	\frac{1}{M}W\mathbf{h} \\
	&=  \lim_{M\rightarrow \infty} \frac{1}{M}\sum_{k=1}^M w_k((w_k^T\mathbf{x})_+ + \epsilon_k), \:\:\:\:\: \epsilon_k \sim \mathcal{N}(0,1) \\
	& = \int_{w\in \mathbb{R}^d} w(w^T\mathbf{x})_+p(w)\:\mathrm{d}w + \lim_{M\rightarrow \infty} \frac{1}{M}\sum_{k=1}^M\epsilon_kw_k \\
	& = \frac{\Sigma\mathbf{x}}{2}
 \end{align*}
 Here, we have omitted the superscript $t$ denoting the iteration number from all variables for ease of presentation. For the first equation, we use the fact that $\mathbf{h}$ has been sampled from normal distribution with unit variance and mean $(\mathbf{x}^Tw_k)_+$. The second equation follows from the law of large numbers and for the third equation we use the fact that $\epsilon_k$ and $w_k$ are independent random variables each with zero mean and the random vector $w$ has been sampled from a distribution with covariance matrix $\Sigma$.
 
 In order to further simplify equation~\eqref{Sigma_update}, we need the values for $\mathbf{h}_0^T\mathbf{h}_0$, $\mathbf{h}_0^T\mathbf{h}_p$ and $\mathbf{h}_p^T\mathbf{h}_p$. However, this is exactly the inner product between the feature representation for the input data, and hence is equivalent to the covariance arc-cosine kernel between the corresponding visible input. Combining all the above results, we get
\begin{equation}
 \begin{split}
 \Sigma^{t+1} =& \Sigma^{t} + \frac{\eta}{2}(\Sigma^{t}\mathbf{x}_0\mathbf{x}_0^T - \Sigma^{t}\mathbf{x}_p\mathbf{x}_p^T + \mathbf{x}_0\mathbf{x}_0^T\Sigma^{t} - \mathbf{x}_p\mathbf{x}_p^T\Sigma^{t}) \\
 				&+ \eta^2(K_{\Sigma^t}(\mathbf{x}_0, \mathbf{x}_0)\mathbf{x}_0\mathbf{x}_0^T + K_{\Sigma^t}(\mathbf{x}_p, \mathbf{x}_p)\mathbf{x}_p\mathbf{x}_p^T \\
 				&- K_{\Sigma^t}(\mathbf{x}_0, \mathbf{x}_p)\mathbf{x}_0\mathbf{x}_p^T - K_{\Sigma^t}(\mathbf{x}_p, \mathbf{x}_0)\mathbf{x}_p\mathbf{x}_0^T   ) 
 \end{split}\label{Sigma_update_final}
 \end{equation}
 Here, $K_{\Sigma^t}$ denotes the covariance arc-cosine kernel function with covariance matrix set to $\Sigma^t$. 
 In order to sample $\mathbf{x}_p$ from $\mathbf{x}_0$, we run a Markov chain with the following transition matrix
 \begin{align*}
 p(\mathbf{x}_{q+1,i} = 1| \mathbf{x}_q; W^t) =& \frac{1}{1+\exp(-{\mathbf{h}_q^TW_i^t})} \\
 							=& \frac{1}{1+\exp(-\frac{\mathbf{x}_q^T\Sigma_i^t}{2})} \:\:,
 \end{align*}
where $\Sigma_i^t$ is the $i^{th}$ column of the covariance matrix and $W_i^t$ is the $i^{th}$ row of matrix $W^t$. It is easy to see that the above Markov chain sampling technique is the same as Gibbs sampling in RBM if the weight matrix is assumed to have infinite number of columns sampled independently from a multivariate Gaussian distribution. The above update gives a stochastic gradient descent method for learning the covariance matrix of the covariance arc-cosine kernel. It is easy to see that after every update the covariance matrix remains positive definite. This is necessary for the resultant kernel to be a valid kernel.

\subsection{Inexact Wide Learning}
Instead of learning the covariance matrix from the data, one can use an RBM to learn the weight matrix $W$. Then assuming the columns of the weight matrix have been sampled from a multivariate Gaussian distribution with zero mean, one can use maximum likelihood to estimate the covariance matrix W. The corresponding covariance matrix is given by
\begin{equation}
\Sigma = \frac{1}{M} WW^T \: ,
\end{equation}
where $W$ is the weight matrix learnt by the RBM and $M$ is the number of columns in $W$.

In our experiments, we found that training the covariance matrix using the first approach took much more time than training an RBM. Secondly, for exact training we had to perform kernel computations and matrix products as mentioned in equation~\eqref{Sigma_update_final} that made each iteration of exact training much slower than the iterations of RBM. It is for this reason, that we use inexact training in the rest of the paper.

\section{DEEP-WIDE LEARNING}
In deep learning, one stacks multiple RBMs one on top of the other such that the output of the previous layer RBM is fed as input to the next layer RBM. Thus, the weight matrix in the second layer is learnt based on the output of the first layer. Similarly, one can stack multiple covariance arc-cosine kernels one on top of the other. However, as mentioned earlier, the feature representation learnt by an arc-cosine kernel has infinite width. Hence, the covariance matrix to be learnt will have infinite number of rows as well as columns. Hence, for learning the covariance matrix in the second layer, one cannot directly use equation~\eqref{Sigma_update_final}. At first glance, it appears that exact learning of the covariance matrix in the second layer is not possible. 

It is here, that kernels come to our rescue. However, when learning multiple layers of kernels notations can become complicated. Hence, it is important to fix the notation. We will use uppercase alphabets to denote both the kernel function and kernel matrices and vectors. 
\begin{enumerate}
\item For a given layer, we use $\tilde{K}$ to denote the inner product between the feature representation of the previous layer. That is, for the first layer, $\tilde{K}(\mathbf{x}, \mathbf{y}) = \mathbf{x}^T\mathbf{y}$. For the second layer $\tilde{K}(\mathbf{x}, \mathbf{y}) = \mathbf{h}(\mathbf{x})^T\mathbf{h}(\mathbf{y})$.
\item For a given layer, we use $\tilde{K}_{\Sigma}$ to denote the inner product between the feature representation of the previous layer using the covariance matrix $\Sigma$. That is, for the first layer, $\tilde{K}_{\Sigma}(\mathbf{x}, \mathbf{y}) = \mathbf{x}^T\Sigma\mathbf{y}$. For the second layer $\tilde{K}(\mathbf{x}, \mathbf{y}) = \mathbf{h}(\mathbf{x})^T\Sigma\mathbf{h}(\mathbf{y})$.
\item For a given layer, we use ${K}_{\Sigma}$ to denote the covariance arc-cosine kernel over the feature representation of the previous layer using the covariance matrix $\Sigma$. .
\end{enumerate}

\subsection{Exact Deep-Wide Learning}
Given the kernel matrix between the feature representation of the previous layer using the covariance matrix $\Sigma$, that is, $\tilde{K}_{\Sigma}$, we compute the covariance arc-cosine kernel matrix $K_\Sigma$ as follows.

\begin{equation}
K_{\Sigma}(\mathbf{x}, \mathbf{y}) = \frac{1}{2\pi}[m(\mathbf{x})m(\mathbf{y})(\sin\theta + (\pi-\theta)\cos\theta) ]\: \label{arc_cosine_from_covariance},
\end{equation}
where 
\begin{align*}
m(\mathbf{x}) &= \tilde{K}_{\Sigma}(\mathbf{x}, \mathbf{x}) \\
m(\mathbf{y}) &= \tilde{K}_{\Sigma}(\mathbf{y}, \mathbf{y}) \\
\theta &= \cos^{-1}\left( \frac{\tilde{K}_{\Sigma}(\mathbf{x}, \mathbf{y})}{\sqrt{\tilde{K}_\Sigma(\mathbf{x}, \mathbf{x}) \tilde{K}_\Sigma(\mathbf{y}, \mathbf{y})} }\right)\enspace .
\end{align*}
In order to compute the covariance arc-cosine kernel matrix over the features learnt by the previous layer, we make use of equation~\eqref{Sigma_update_final}. Let $\mathbf{h}(a)$ and $\mathbf{h}(b)$ be the infinite dimensional feature representation learnt by the previous layer. We pre-multiply equation~\eqref{Sigma_update_final} by $\mathbf{h}(a)^T$ and post-multiply it by $\mathbf{h}(b)$ to get 
\begin{equation}
\begin{split}
&\tilde{K}_{\Sigma^{t+1}}(a,b) = \tilde{K}_{\Sigma^t}(a,b) \\ 
				&+ \frac{\eta}{2}(\tilde{K}_{\Sigma^t}(a,\mathbf{x}_0)\tilde{K}(\mathbf{x}_0,b) - \tilde{K}_{\Sigma^t}(a,\mathbf{x}_p)\tilde{K}(\mathbf{x}_p, b)) \\
				& + \frac{\eta}{2}(\tilde{K}(a,\mathbf{x}_0)\tilde{K}_{\Sigma^t}(\mathbf{x}_0,b) - \tilde{K}(a,\mathbf{x}_p)\tilde{K}_{\Sigma^t}(\mathbf{x}_p,b)) \\
 				&+ \eta^2 K_{\Sigma^t}(\mathbf{x}_0, \mathbf{x}_0)\tilde{K}(a, \mathbf{x}_0)\tilde{K}(\mathbf{x}_0,b) \\ 
 				&+ \eta^2 K_{\Sigma^t}(\mathbf{x}_p, \mathbf{x}_p)\tilde{K}(a, \mathbf{x}_p)\tilde{K}(\mathbf{x}_p,b) \\
 				&- \eta^2 K_{\Sigma^t}(\mathbf{x}_0, \mathbf{x}_p)\tilde{K}(a, \mathbf{x}_0)\tilde{K}(\mathbf{x}_p,b) \\ 
 				&- \eta^2 K_{\Sigma^t}(\mathbf{x}_p, \mathbf{x}_0)\tilde{K}(a, \mathbf{x}_p)\tilde{K}(\mathbf{x}_0,b) \enspace .
\end{split} \label{kernel_update}
\end{equation}

The above equation updates each entry of the kernel matrix in a sequential fashion. However, one can choose to update the entire kernel matrix in one go by using the following equation.
\begin{equation}
\begin{split}
\tilde{K}_{\Sigma^{t+1}} =& \tilde{K}_{\Sigma^t} + \frac{\eta}{2}(\tilde{K}_{\Sigma^{t}, \mathbf{x}_0}\tilde{K}_{\mathbf{x}_0}^T - \tilde{K}_{\Sigma^{t}, \mathbf{x}_p}\tilde{K}_{\mathbf{x}_p}^T) \\
							&+\frac{\eta}{2}(\tilde{K}_{\mathbf{x}_0}\tilde{K}_{\Sigma^{t}, \mathbf{x}_0}^T - \tilde{K}_{\mathbf{x}_p}\tilde{K}_{\Sigma^{t}, \mathbf{x}_p}^T) \\
							&+\eta^2{K}_{\Sigma^t}(\mathbf{x}_0, \mathbf{x}_0)\tilde{K}_{\mathbf{x}_0}\tilde{K}_{\mathbf{x}_0}^T \\
							&+\eta^2{K}_{\Sigma^t}(\mathbf{x}_p, \mathbf{x}_p)\tilde{K}_{\mathbf{x}_p}\tilde{K}_{\mathbf{x}_p}^T \\
							&-\eta^2{K}_{\Sigma^t}(\mathbf{x}_0, \mathbf{x}_p)\tilde{K}_{\mathbf{x}_0}\tilde{K}_{\mathbf{x}_p}^T \\
							&-\eta^2{K}_{\Sigma^t}(\mathbf{x}_p, \mathbf{x}_0)\tilde{K}_{\mathbf{x}_p}\tilde{K}_{\mathbf{x}_0}^T \enspace ,
\end{split} \label{kernel_matrix_update}
\end{equation}
where $\tilde{K}$ and $\tilde{K}_{\Sigma^t}$ denote the kernel matrices corresponding to the kernel functions $\tilde{K}$ and $\tilde{K}_{\Sigma^t}$ respectively as defined above and $\tilde{K}_{\mathbf{x}_0}$ and $\tilde{K}_{\Sigma^{t}, \mathbf{x}_0}$ denote the column in kernel matrices $\tilde{K}$ and $\tilde{K}_{\Sigma_t}$ corresponding to $\mathbf{x}_0$ respectively. It is interesting to note the similarity between the above equation and equation~\eqref{Sigma_update_final}.

This suggests the following steps for computing the kernel matrix in the second layer based on the kernel matrix of the previous layer.
\begin{enumerate}
\item Let $\tilde{K}$ be the kernel matrix corresponding to the first layer. Initialize $\tilde{K}_{\Sigma^0}$ to be a random positive definite matrix of size $N\times N$.
\item Update the kernel matrix $\tilde{K}_{\Sigma^t}$ using equation~\eqref{kernel_matrix_update} until convergence.
\item The kernel matrix $K$ of the second layer can then be computed from $\tilde{K}_{\Sigma^t}$ by composing the arc-cosine kernel with the kernel matrix $K_{\Sigma^t}$ as given in equation \eqref{arc_cosine_from_covariance}. 
\end{enumerate}

\subsection{Inexact Deep-Wide Learning}
Exact learning of the kernel matrix as given above requires the entire matrix to be present in memory. Since unsupervised feature learning only works when the number of instances is huge, this means that the kernel matrix will also be very huge. Hence, learning using such a huge kernel matrix will be infeasible both in terms of memory and processing time requirements.

Hence, we tried inexact approaches to extend the architecture to multiple layers. In the first approach, we learn a finite dimensional first layer using RBM. Next, a covariance arc-cosine kernel is learnt on top of the activities of the first level RBM as mentioned in the previous section. However, we found that for all datasets that we tried, this approach resulted in reduction in accuracy. For instance, for MNIST digit recognition task, the accuracy reduced from $99.05\%$ to $97.85\%$.

We also tried to sample a subset of features by applying kernel PCA for the covariance arc-cosine kernel. However, this method further resulted in reduction in performance. Hence, for the rest of the experiments, only the first layer is learnt using RBM.

\section{DISCUSSION}
In order to understand why a covariance arc-cosine kernel should work, we divert ourselves from feature learning to feature extraction. It has been known for a long time now that gradient~(orientation as well as magnitude) at every pixel location is a more natural representation for an image then the actual pixel values. For instance, in SIFT~\cite{dalal2005histograms}, the gradient orientation at each pixel is binned into one of the finitely many bins available to get a finite dimensional vector representation. In soft binning, multiple bins are allowed to be non-zero. The feature representation of a patch is then computed by doing a weighted sum of the gradient orientation vector at all pixels in the patch. Finally a linear classifier is applied on the resultant feature representation.

The above model is equivalent to defining a linear kernel on the weighted sum of binned gradient orientation vectors. However, instead of computing a linear kernel between the binned orientation vectors, one can choose to compute an RBF kernel over the gradient orientation vectors themselves~(without binning). This is equivalent to computing an inner product between the infinite dimensional representation of the orientation vectors. As shown in~\cite{bo2010kernel}, this very small trick results in improved performance.

Here, our idea is very similar. When we learn the weight matrix $W$ using an RBM and apply the resultant matrix on visible data, we get a finite dimensional representation of the data. If we assume now that each bin has been labelled by a column $w_j$ of the weight matrix $W$, then this approach is equivalent to soft binning, where $j^{th}$ bin is non-zero, if and only if $w_j^T\mathbf{x}$ is positive~(assuming rectified linear units). This creates a finite dimensional vector representation of the data.

However, if we learn the distribution of the columns in the weight matrix $W$, we can, in principle, project the data to infinite dimensions by sampling infinitely many vectors $w_j$ from the distribution, using the kernel trick. A covariance arc-cosine kernel makes the additional assumption that the distribution of the columns in $W$, is multivariate Gaussian. Hence, while an RBM can only bin the data into finite many bins, use of covariance arc-cosine kernel allows one to bin the data in infinite number of bins by using the kernel trick. This is also a reason why we term our proposed learning method as wide learning. An important point to note at this juncture is that though the final model has infinite width, the model learnt by RBM still has a finite small width. Hence, the number of parameters in the model are much lesser than in a deep learning model. This approach is very fast since only a single RBM is trained and no fine-tuning needs to be done.

\section{EXPERIMENTS}
We tested the covariance kernel so obtained for many datasets commonly used for comparing deep learning architectures.
\subsection{MNIST}
The MNIST dataset~\cite{lecun1998gradient} consists of grayscale images of handwritten digits from 0 to 9 with $50,000$ training and $10,000$ test examples. We normalize the pixel values to be between $0$ and $1$. Except that, we do not use any preprocessing for the dataset. A standard Bernoulli RBM with $1000$ hidden units is trained on the raw pixel values using stochastic gradient descent with a fixed momentum of 0.5. In our experiments for MNIST dataset, we found that fixing the bias to zero doesn't affect the performance of the final model. 

Finally, we use the weight matrix $W$ learnt by RBM to compute the kernel matrix. The kernel matrix is fed into a regularized least square kernel classifier. We use one-vs-one classification to classify the MNIST images. Using a single layer, we were able to achieve an error rate of $0.95\%$ after $20$ epochs, which is quite better compared to the  $1.25\%$ error rate obtained by a deep belief network~\cite{hinton2006fast}. This is surprising since a deep belief network uses multiple stacked RBMs, while we used a single RBM in our model. Furthermore, both the RBMs were trained similarly. This suggests that a single RBM might be enough to learn the invariances in the data.

In order to show the advantage of having a representation of infinite width (that is, by using the kernel), we compare the performance of the covariance arc-cosine kernel against a neural network with 1 hidden layer with number of epochs of training in Figure~\ref{AccuracyVsEpoch}.  The network has $1000$ hidden units. The parameters of both the neural network and the covariance arc-cosine kernel are obtained using a restricted Boltzmann machines with $1000$ hidden units. No fine tuning is done for either of the models. There are two important observations that can be made from the figure. Firstly, the infinite width model reaches an acceptable performance in a very few number of epochs. Secondly, even after running the unsupervised learning algorithm for 20 epochs, the accuracy of the finite width model never comes any close to the accuracy of the infinite width model after 2 epochs. In fact, when we ran the training algorithm for 300 epochs, the accuracy of the finite width model converged to $97.93\%$. This is in stark contrast with the accuracy achieved by the infinite width model after 1 epoch~($98.9\%$). This is a very surprising result, which suggests that even if the training time for deep learning is small, a covariance arc-cosine kernel can still give acceptable performance. 

Finally, we will briefly mention about the time required for training and testing for finite and infinite width models compared in the previous paragraph. The unsupervised learning phase in both the models comprises of learning a weight matrix from the data by defining a restricted Boltzmann machine. In infinite width model, the weight matrix is used to learn a covariance matrix $\Sigma$. The covariance arc-cosine kernel is then computed for the data using the covariance matrix. Finally, an algorithm based on kernel methods (such as SVM), is then used to compute the parameters for the last layer. On the other hand, for the finite width model, a linear classifier is learnt on top of the activities of the hidden layer. 
Thus, the only difference in computation cost for finite and infinite width model lies in time taken to learn the last layer. Kernel methods can be more computationally costly, when the number of instances is huge, because of their quadratic~(at least) dependence on the number of instances.

Two variants of MNIST are also considered for comparison as listed below:
\begin{figure}
\vspace{.3in}
\includegraphics[width = 7cm]{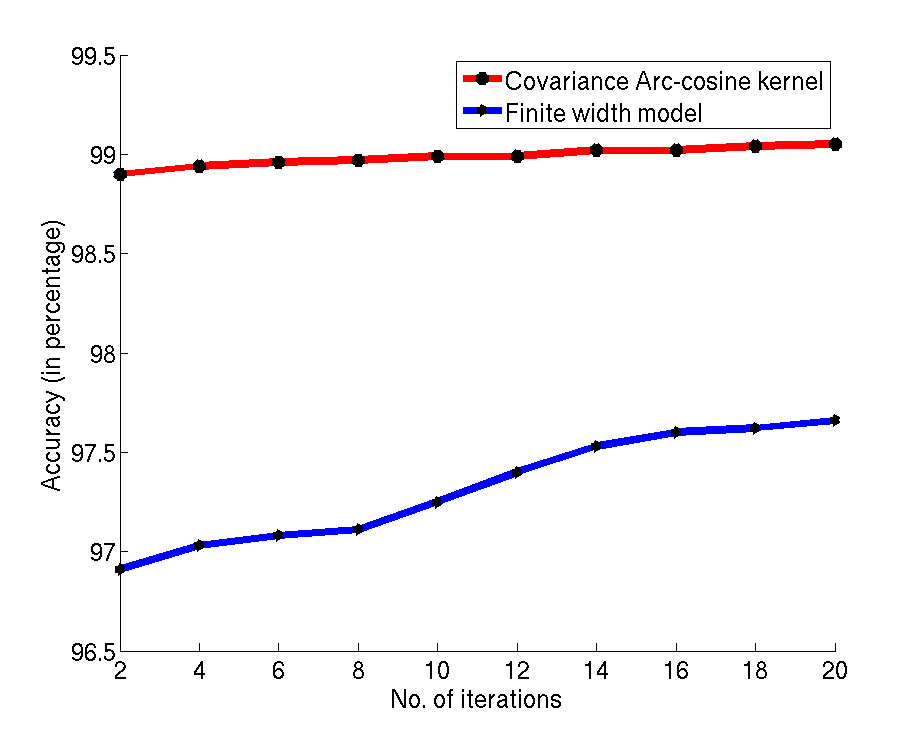}
\vspace{.3in}
\caption{Comparison of finite width model learnt by an RBM (not fine-tuned) against the infinite width model obtained by a covariance arc-cosine kernel. An infinite width model gives acceptable performance~(around 98.9\% accuracy) even after 2 epochs.}
\label{AccuracyVsEpoch}
\end{figure}


\begin{table*}[ht]
\caption{Comparison of covariance arc-cosine kernel learnt by our model against other classifiers. Some of the results have been borrowed from~\cite{larochelle2007empirical}, but only the best results are kept. The best results for each dataset is indicated in bold. The numbers in parenthesis indicate the number of unlearnt layers in the model. The default number of unlearnt layers is 0.} 
\centering 
\tabcolsep=0.11cm
\begin{tabular}{c c c c c c} 
\hline\hline 
 & MNIST & MNIST & MNIST & Rectangles & Convex\\ [0.5ex] 
 		   & original & rotated & random backgd. \\
\hline 
SVM-rbf &1.6\% &10.38\%  &14.58\% &2.15\% &19.13\% \\
SVM-poly~\cite{decoste2002training} &1.22\%  &13.61\%  &16.62\% &2.15\% &19.82\% \\ 
Neural network~(one hidden layer) &1.9\% &17.62\% &20.04\% &7.16\% &32.25\%  \\
DBN~\cite{hinton2006fast}   & 1.25\% &12.30\% &\textbf{6.73}\% & 2.60\% &18.63\% \\
Stacked Autoencoder~\cite{bengio2007greedy} &1.4\% &11.43\%  & 11.28\% &2.41\% &18.41\% \\
Arc-cosine kernel~\cite{cho2010large} & 1.38\%(5) &11.22\%(5) & 16.14\%(5) &2.27\%(15)  &17.15\%(5)  \\
\textbf{Covariance arc-cosine kernel}& \textbf{0.95}\% & \textbf{8.11}\% &8.9\% & \textbf{1.49}\%(25) & \textbf{17.02}\% \\ [1ex] 
\hline 
\end{tabular}
\label{table:nonlin} 
\end{table*}

\begin{enumerate}
\item{\bf Rotated MNIST data set:}
This is a modified version of MNIST dataset, where the digits are rotated by an angle generated uniformly between $0$ and $2\pi$. This dataset is one of the several MNIST variations generated for comparing deep vs shallow architectures in~\cite{larochelle2007empirical}. The dataset has $12,000$ training samples and $50,000$ test samples. 

\item{\bf MNIST with random background:}
This is also a modified version of the MNIST dataset where the background pixels have been sampled randomly between $0$ and $255$. This dataset also has $12,000$ training and $50,000$ test samples.
\end{enumerate}

\subsection{Tall and wide rectangles}
This dataset~\cite{larochelle2007empirical} consists of rectangles of varying widths and heights. The aim is to classify the rectangles into classes, where the rectangles in one class have more height than width, while the rectangles in the other class have more width than height. We trained an RBM over the pixel values and computed the covariance matrix from the learnt weight matrix $W$. This covariance matrix was used in the covariance arc-cosine kernel in the first layer. 

For this dataset, we found that using a first layer of covariance arc-cosine kernel, followed by multiple layers of arc-cosine kernels where the covariance matrix is set to identity, resulted in improvement in performance. Note that, this is still not a case of deep learning since only a single layer is learnt. In fact, this approach can be shown as equivalent to learning the covariance matrix of a different kernel
which is obtained by composing multiple arc-cosine kernels with identity covariance matrix. For more details, we encourage the reader to refer to~\cite{cho2010large}, where composition of arc-cosine kernels is discussed in further detail. The result for covariance arc-cosine kernel against other standard classifiers is given in~Table~\ref{table:nonlin}. Clearly, the best results are obtained when the first layer is learnt using a covariance arc-cosine kernel.

This dataset~\cite{larochelle2007empirical} consists of black and white images of convex and concave sets. 
The task is to separate the concave sets from convex sets. The results are given in Table~\ref{table:nonlin}.

\section{Conclusion}
In this paper, we proposed the notion of wide learning, that makes the fine-tuning stage commonly used in deep architectures redundant. We have given exact as well as inexact methods for learning in such models. We found that for the datasets considered, whenever we replace a finite width layer by a layer of infinite width, this results in drastic improvement in performance. Furthermore, use of a single layer severely reduces the number of hyper-parameters to be estimated, thereby saving time in computationally costly grid searches. 
Further experimentation on more complicated datasets, such as natural image patches, is needed to test its suitability for general AI tasks.
\bibliographystyle{apalike}
\bibliography{deep}

\begin{thebibliography}{}

\bibitem[Bach et~al., 2004]{bach2004multiple}
Bach, F.~R., Lanckriet, G.~R., and Jordan, M.~I. (2004).
\newblock Multiple kernel learning, conic duality, and the {SMO} algorithm.
\newblock In {\em Proceedings of the 21st {I}nternational {C}onference on
  Machine learning}, page~6. ACM.

\bibitem[Bengio, 2009]{bengio2009learning}
Bengio, Y. (2009).
\newblock Learning deep architectures for {AI}.
\newblock {\em Foundations and trends{\textregistered} in Machine Learning},
  2(1):1--127.

\bibitem[Bengio et~al., 2007]{bengio2007greedy}
Bengio, Y., Lamblin, P., Popovici, D., and Larochelle, H. (2007).
\newblock Greedy layer-wise training of deep networks.
\newblock {\em Advances in Neural Information Processing Systems}, 19:153.

\bibitem[Bergstra et~al., 2011]{bergstra2011algorithms}
Bergstra, J., Bardenet, R., Bengio, Y., K{\'e}gl, B., et~al. (2011).
\newblock Algorithms for hyper-parameter optimization.
\newblock In {\em Advances in Neural Information Processing Systems (NIPS
  2011)}.

\bibitem[Blitzer et~al., 2005]{blitzer2005distance}
Blitzer, J., Weinberger, K.~Q., and Saul, L.~K. (2005).
\newblock Distance metric learning for large margin nearest neighbor
  classification.
\newblock In {\em Advances in Neural Information Processing Systems}, pages
  1473--1480.

\bibitem[Bo et~al., 2010]{bo2010kernel}
Bo, L., Ren, X., and Fox, D. (2010).
\newblock Kernel descriptors for visual recognition.
\newblock In {\em Advances in Neural Information Processing Systems}, pages
  244--252.

\bibitem[Cho and Saul, 2010]{cho2010large}
Cho, Y. and Saul, L.~K. (2010).
\newblock Large-margin classification in infinite neural networks.
\newblock {\em Neural Computation}, 22(10):2678--2697.

\bibitem[Ciresan et~al., 2012]{ciresan2012multi}
Ciresan, D., Meier, U., and Schmidhuber, J. (2012).
\newblock Multi-column deep neural networks for image classification.
\newblock In {\em Computer Vision and Pattern Recognition (CVPR)}, pages
  3642--3649. IEEE.

\bibitem[Dalal and Triggs, 2005]{dalal2005histograms}
Dalal, N. and Triggs, B. (2005).
\newblock Histograms of oriented gradients for human detection.
\newblock In {\em Proceedings of IEEE Conference on Computer Vision and Pattern
  Recognition (CVPR)}, volume~1, pages 886--893. IEEE.

\bibitem[Decoste and Sch{\"o}lkopf, 2002]{decoste2002training}
Decoste, D. and Sch{\"o}lkopf, B. (2002).
\newblock Training invariant support vector machines.
\newblock {\em Machine Learning}, 46(1-3):161--190.

\bibitem[Fischer and Igel, 2012]{fischer2012introduction}
Fischer, A. and Igel, C. (2012).
\newblock An introduction to restricted {B}oltzmann machines.
\newblock In {\em Progress in Pattern Recognition, Image Analysis, Computer
  Vision, and Applications}, pages 14--36. Springer.

\bibitem[Hinton, 2002]{hinton2002training}
Hinton, G.~E. (2002).
\newblock Training products of experts by minimizing contrastive divergence.
\newblock {\em Neural Computation}, 14(8):1771--1800.

\bibitem[Hinton et~al., 2006]{hinton2006fast}
Hinton, G.~E., Osindero, S., and Teh, Y.-W. (2006).
\newblock A fast learning algorithm for deep belief nets.
\newblock {\em Neural {C}omputation}, 18(7):1527--1554.

\bibitem[Huang et~al., 2004]{huang2004extreme}
Huang, G.-B., Zhu, Q.-Y., and Siew, C.-K. (2004).
\newblock Extreme learning machine: a new learning scheme of feedforward neural
  networks.
\newblock In {\em International Joint Conference on Neural Networks}, volume~2,
  pages 985--990. IEEE.

\bibitem[Krizhevsky et~al., 2012]{krizhevsky2012imagenet}
Krizhevsky, A., Sutskever, I., and Hinton, G.~E. (2012).
\newblock Imagenet classification with deep convolutional neural networks.
\newblock In {\em NIPS}, volume~1, page~4.

\bibitem[Lamblin and Bengio, 2010]{lamblin2010important}
Lamblin, P. and Bengio, Y. (2010).
\newblock Important gains from supervised fine-tuning of deep architectures on
  large labeled sets.
\newblock In {\em NIPS* 2010 Deep Learning and Unsupervised Feature Learning
  Workshop}.

\bibitem[Larochelle et~al., 2007]{larochelle2007empirical}
Larochelle, H., Erhan, D., Courville, A., Bergstra, J., and Bengio, Y. (2007).
\newblock An empirical evaluation of deep architectures on problems with many
  factors of variation.
\newblock In {\em Proceedings of the 24th International Conference on Machine
  Learning}, pages 473--480. ACM.

\bibitem[LeCun et~al., 1998]{lecun1998gradient}
LeCun, Y., Bottou, L., Bengio, Y., and Haffner, P. (1998).
\newblock Gradient-based learning applied to document recognition.
\newblock {\em Proceedings of the IEEE}, 86(11):2278--2324.

\bibitem[Lowe, 2004]{lowe2004distinctive}
Lowe, D.~G. (2004).
\newblock Distinctive image features from scale-invariant keypoints.
\newblock {\em International Journal of Computer Vision}, 60(2):91--110.

\bibitem[Nair and Hinton, 2010]{nair2010rectified}
Nair, V. and Hinton, G.~E. (2010).
\newblock Rectified linear units improve restricted boltzmann machines.
\newblock In {\em Proceedings of the 27th International Conference on Machine
  Learning (ICML-10)}, pages 807--814.

\bibitem[Utgoff and Stracuzzi, 2002]{utgoff2002many}
Utgoff, P.~E. and Stracuzzi, D.~J. (2002).
\newblock Many-layered learning.
\newblock {\em Neural {C}omputation}, 14(10):2497--2529.

\bibitem[Wan et~al., 2013]{wan2013regularization}
Wan, L., Zeiler, M., Zhang, S., Cun, Y.~L., and Fergus, R. (2013).
\newblock Regularization of neural networks using dropconnect.
\newblock In {\em Proceedings of the 30th International Conference on Machine
  Learning (ICML-13)}, pages 1058--1066.

\bibitem[Xiao et~al., 2010]{xiao2010sun}
Xiao, J., Hays, J., Ehinger, K.~A., Oliva, A., and Torralba, A. (2010).
\newblock Sun database: Large-scale scene recognition from abbey to zoo.
\newblock In {\em Proceedings of IEEE conference on Computer Vision and Pattern
  Recognition (CVPR)}, pages 3485--3492. IEEE.

\end{thebibliography}
\end{document}